\documentclass{article}

\usepackage[final,nonatbib]{neurips_2019}

\usepackage[utf8]{inputenc} % allow utf-8 input
\usepackage[T1]{fontenc}    % use 8-bit T1 fonts
\usepackage{hyperref}       % hyperlinks
\usepackage{url}            % simple URL typesetting
\usepackage{booktabs}       % professional-quality tables
\usepackage{amsfonts}       % blackboard math symbols
\usepackage{nicefrac}       % compact symbols for 1/2, etc.
\usepackage{microtype}      % microtypography
\usepackage{xcolor}
\usepackage{graphicx}
\usepackage{caption}
\usepackage{subcaption}

\title{Inundation Modeling in Data Scarce Regions}
\author{
  Zvika Ben-Haim \hspace{1em}
  Vladimir Anisimov \hspace{1em}
  Aaron Yonas \hspace{1em}
  Varun Gulshan
  \vspace{.5em} \\
  \textbf{Yusef Shafi} \hspace{1em}
  \textbf{Stephan Hoyer} \hspace{1em}
  \textbf{Sella Nevo} \hspace{1em}
  \vspace{.5em} \\ 
  Google Research \\
  \{zvika, vovani, aaronyonas, varungulshan, yusef, shoyer, sellanevo\}@google.com
}

\begin{document}

\maketitle

\begin{abstract}
  Flood forecasts are crucial for effective individual and governmental protective action. The vast majority of flood-related casualties occur in developing countries, where providing spatially accurate forecasts is a challenge due to scarcity of data and lack of funding. This paper describes an operational system providing flood extent forecast maps covering several flood-prone regions in India, with the goal of being sufficiently scalable and cost-efficient to facilitate the establishment of effective flood forecasting systems globally.
\end{abstract}

\section{Introduction}

Floods are among the most common and most deadly natural disasters in the world, affecting hundreds of millions of people and causing thousands of fatalities annually \cite{doocy2013human}. Flood forecasting systems are an effective mitigation tool, reducing fatalities and economic costs by about a third \cite{malilay1997floods}.

The spatial accuracy of alerts is vital both in allowing governments to plan effective mitigation and relief efforts, and in providing actionable information to individuals. Unfortunately, the vast majority of those exposed to flooding, and an even larger portion of fatalities, are located in regions that do not have access to spatially accurate forecasting systems \cite{hirpa2016saving}.

This paper describes an operational system providing high-resolution, high-accuracy flood forecasts for riverine floods. The system covers an area of 11,600 km$^2$ along the Ganges, Brahmaputra, and Ghaghara rivers in India. Throughout the 2019 monsoon season (July--October), the system generated over 170 inundation forecast maps for multiple, continuously-evolving severe floods. These maps were delivered in real time, both to government agencies (through dedicated communication channels) and to individuals (via Android notifications and other Google products).

At a high level, our system receives a measurement or forecast of river water level, and predicts the extent of the resulting flood. The system consists of the following components. First, we generate high-resolution elevation maps. These are critical for the accuracy of any inundation model, but producing these globally and cost-effectively is challenging. Second, we obtain real-time measurements and forecasts of river water levels, measured either using a stream gauge or a hydrological model. We currently rely on government agencies to obtain these data. Third, we perform hydraulic modeling of the river and floodplain, combining physics-based and ML-based solutions to obtain sufficient prediction accuracy. Finally, we disseminate the forecasts to the affected population. To simplify interpretation of the forecast, we generate a map containing three regions, representing three levels of risk: ``Some risk,'' ``Higher risk,'' and ``Highest risk.''

The contents of this paper describe our current operational system, as well as concrete efforts we are actively working on. In the process of scaling our system, we will need to tackle new climates, ground conditions, and other obstacles that will require additional innovations. Nevertheless, we believe the concepts presented in this paper are first steps towards a truly planet-scale system, which is required to address the thousands of unnecessary deaths floods cause every year.

\section{Digital elevation model generation} \label{dem_generation}

Lack of a publicly accessible, global, high resolution digital elevation model (DEM)\footnote{In this paper, digital surface models (DSMs) include groundcover such as trees and buildings; digital terrain models (DTMs) approximate ``bare earth,'' with groundcover removed; and DEM is an umbrella term that encompasses both.} is a critical limiting factor in flood modeling. The most widely used DEMs are NASA's SRTM \cite{srtm} and JAXA's AW3D30 \cite{aw3d30}, both of which are freely available at 30 meter horizontal resolution.  These DEMs have significant vertical error \cite{yamazaki} and are aging: the data for SRTM and AW3D30 was gathered in 2000 and during 2006--2011, respectively.

Despite their limitations, these DEMs are used by hydrologists because the commercial alternatives are prohibitively expensive. We have been unable to build accurate flood models using publicly accessible DEMs. Instead, we repurpose existing satellite images to construct high-resolution DEMs.

\subsection{DEM generation challenges}

Reconstructing DEMs from general-purpose satellite images is a particularly challenging variation of the photogrammetric bundle adjustment problem \cite{triggs}, where the corpus is:

\begin{itemize}
\item
Multi-sensor: We fuse data from many satellites with different image resolutions, frequency response curves, types of optical distortion, and spatial errors.

\item
Multitemporal: Our images were captured over months or years. This makes registration challenging, as the images may differ substantially, e.g., due to seasonal variations. Furthermore, the topography itself changes over time (e.g., due to riverine erosion), and consequently older images must sometimes be discarded in favor of newer ones.

\item
Ad hoc: To create DEMs, operators carefully plan the rotational motion of the satellite to obtain multiple ``stereo'' views of the target from different angles. In our case, the viewing directions are effectively random.
\end{itemize}

\subsection{Camera model correction and DSM generation}

Our system first gathers all satellite images that may intersect a given region. Then, in a coarse-to-fine manner, it simultaneously solves for surface elevations and adjustments to the images' camera models. Due to the diversity of satellite inputs, we have found it more robust to use dense digital surface model (DSM) estimation combined with patch-based cross-correlation alignment than to use the more typical sparse 3D estimation from feature matching. Each correlated pair of patches provides a local translation. Combining many local translations in an iterative solver leads to general spatially-varying alignment fields.  The cross-correlation process generates confidence values that allow for spatially-adaptive regularization of the alignment fields.

We then hold the camera models fixed and use standard multi-view stereo methods to estimate a depth map for each image.  To create the DSM, we merge the depth maps by determining, at each horizontal location, the elevation yielding the greatest agreement.  Of course, some depth maps may disagree due to changes over time, clouds, etc.  We use a graph-cut solver to find the best consensus elevation.  Both the stereo and depth map merging steps are performed in a coarse-to-fine manner for computational efficiency.

\subsection{DTM generation}

Since water flows around trees, digital surface models (DSMs) can distort hydraulic simulations.  For example, a row of trees planted as a windbreak can, in a simulation, erroneously behave like a dam or levee.  One way to create a bare earth digital terrain model (DTM) is to start with a DSM (plus projected images) and train a convolutional neural network or other ML model to identify where the DSM and DTM differ.  Trees and buildings may then be removed and filled with a smoothly interpolated estimate of the ground.  Of course, in the case of large, continuous forests, additional information, such as canopy-penetrating radar or tree-height surveys, would be needed to accurately pick the right ground elevation.  The ideal solution for flood modeling is a hybrid between a DSM and DTM that excludes trees but retains the man-made structures that can affect water flow.

\subsection{River segmentation}
\label{section:river_segment}

\begin{figure}[ht]
\begin{subfigure}[b]{0.45\textwidth}
  \centering
  \includegraphics[width=\textwidth]{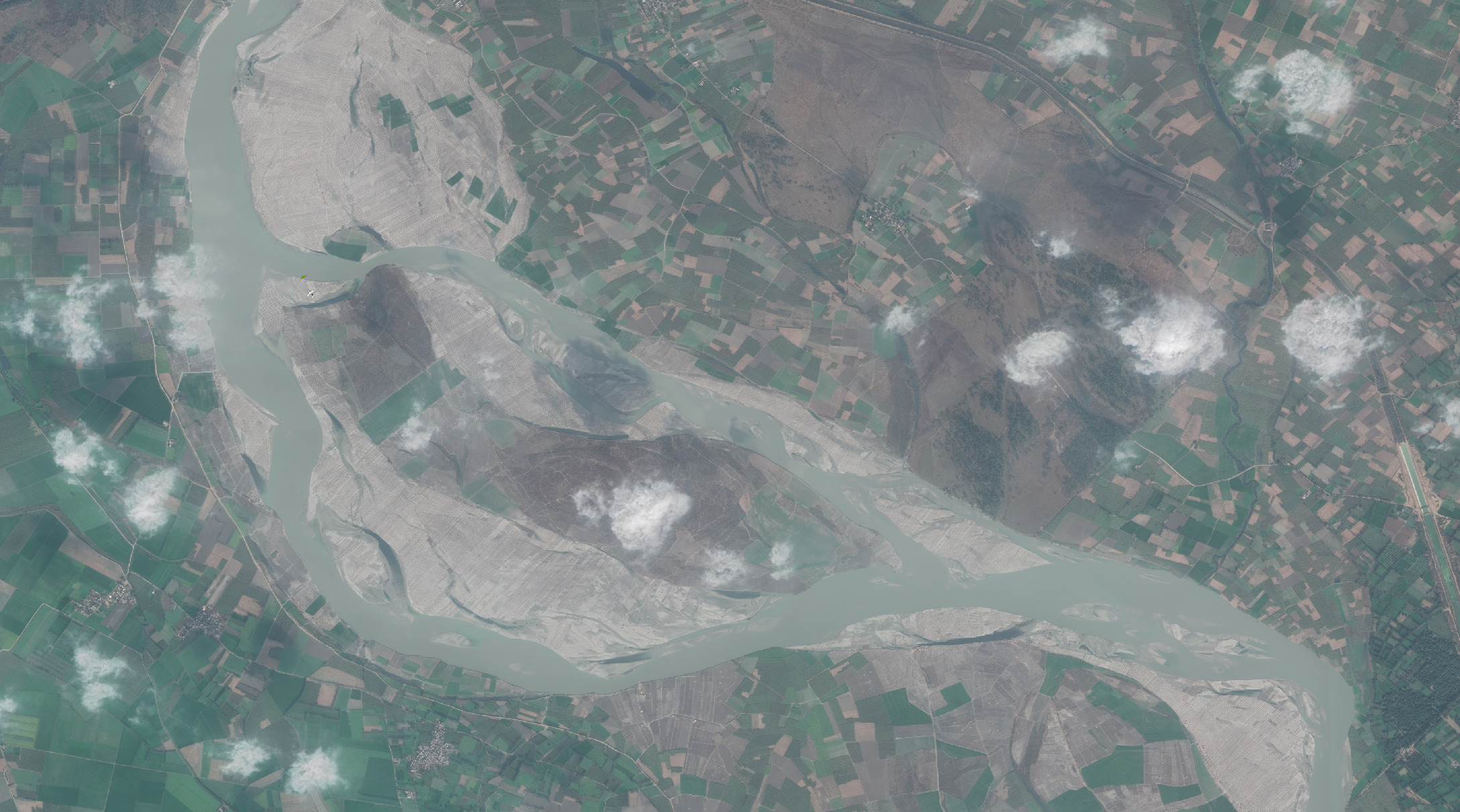} 
%  \caption{Satellite image from test region}
\end{subfigure}
\hfill
\begin{subfigure}[b]{0.45\textwidth}
  \centering
  \includegraphics[width=\textwidth]{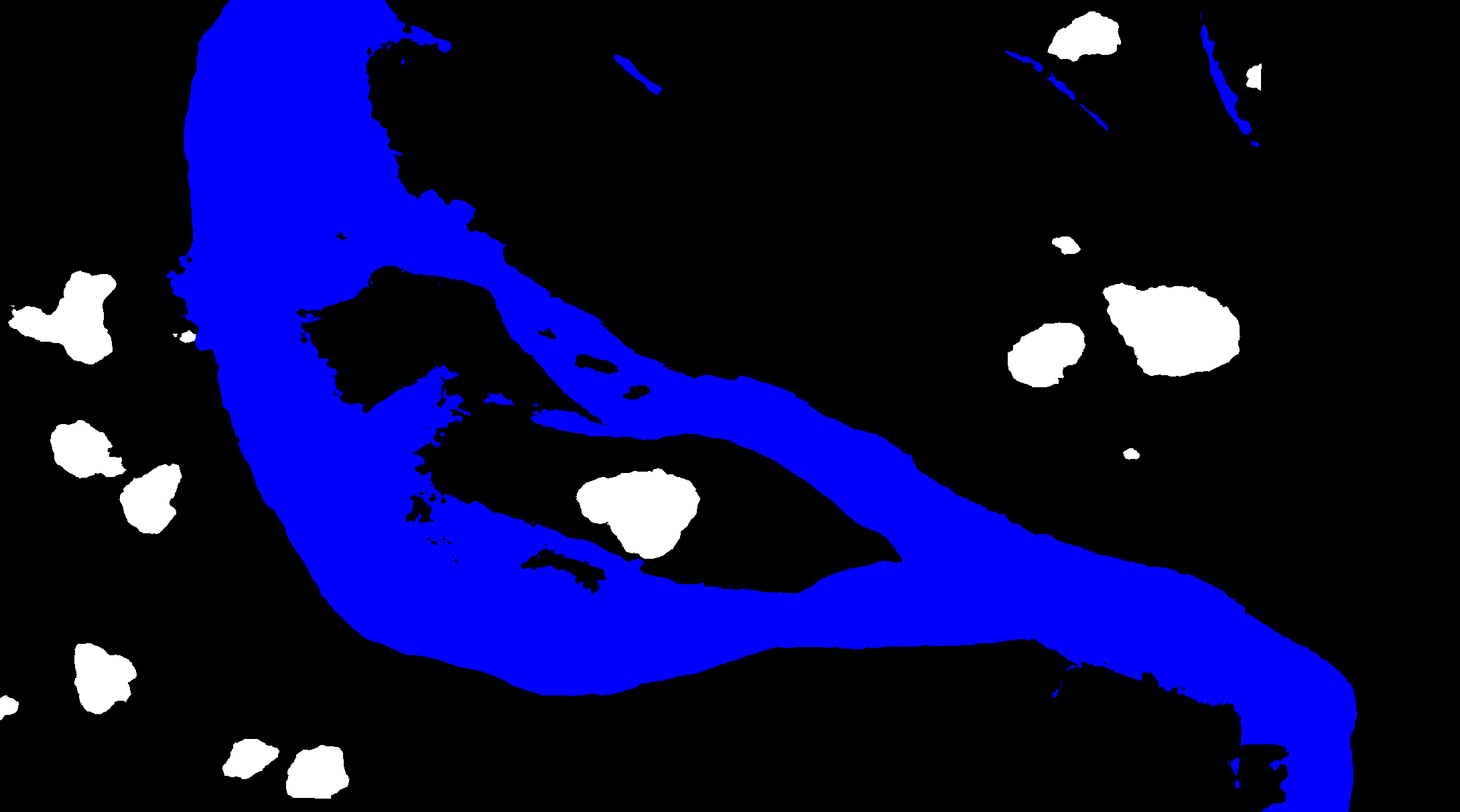} 
%  \caption{Riverbed (blue) and cloud (white) mask}
\end{subfigure}
\caption{Satellite image of a test region (left), and corresponding inferred labels for the image (right). Blue represents riverbed and white represents clouds.}
\label{fig:riverbedmask}
\end{figure}

Because images cannot be correlated and aligned over water, we segment river pixels and replace the river DEM with a constant-slope riverbed.   Manually creating this riverbed mask for large areas is time-consuming and expensive, and we are working to automate this using ML-based segmentation. In order to create training data for this task, we have trained humans to mark out regions in a satellite image corresponding to riverbed and clouds. The images are labeled at a resolution of $4m \times 4m$ per pixel. We currently have $3.8 \times 10^9$ labeled pixels, which are used for training and validation. The DeepLabv3+ \cite{chen2018encoder} model is trained to segment the riverbed and clouds using this data. A pre-initialized Xception network \cite{chollet2017xception} trained on the COCO dataset \cite{coco2014} is used as the encoder backbone of the model. Our early qualitative results look promising (see Figure~\ref{fig:riverbedmask}), and we are working towards quantifying these metrics before using this system in production.

\section{Inundation forecasting}

\subsection{Hydraulic modeling} \label{section:hydraulic}

Hydraulic models simulate the flow of water in a given topography. When properly calibrated, hydraulic models are capable of accurately predicting inundation extent and water depth. The classical approach is to time-integrate the Saint-Venant shallow water partial differential equations (PDEs). Ignoring the negligible advection term, the PDEs are given by \cite{vreugdenhil94, de2012improving}
\begin{equation} \label{eq:st-venant}
\frac{\partial h}{\partial t} + \frac{\partial q_x}{\partial x} + \frac{\partial q_y}{\partial y} = 0
\quad \textrm{ and } \quad
\frac{\partial q_i}{\partial t} + g h \frac{\partial (h+z)}{\partial i} +
\frac{g n^2}{h^{7/3}} \|\mathbf{q}\| q_i = 0, \quad i \in \{ x, y \}
\end{equation}
where $\mathbf{q} = (q_x, q_y)$ is the flux, $h$ is the water height, $z$ is the surface elevation (topography), $g$ is the gravitational acceleration constant, and $n$ is the Manning friction coefficient. The first equation in (\ref{eq:st-venant}) represents conservation of mass: when water flows into a cell, the water height rises. The second equation represents Newton's second law: water accelerates (its flux increases) due to gravity when there are differences in water height, and decelerates due to friction.

Our implementation is based on the numerical scheme of \cite{de2012improving} for solving the inertial form of the Saint-Venant equations. The region of interest is represented as a 2D grid with cell sizes around 10~m.

To integrate (\ref{eq:st-venant}), one must provide boundary conditions, including the incoming discharge. Unfortunately, discharge is difficult to measure directly, and such measurements are not available in real-time for most regions in India. Instead, we rely only on river \emph{water level} measurements and forecasts, which are more commonly available.

To utilize river water level forecasts, we assume that the river level changes slowly, so that the flood extent at any given moment is approximated by a simulation with constant influx at $t \rightarrow \infty$. This is a good approximation for large rivers, which are the current focus of our floods warning system: In such rivers, water level changes relatively slowly due to the large amounts of water involved. We select numerous different boundary conditions and simulate each until it approaches a steady state. The inundation extent at steady state is recorded and indexed by the simulated water level at the stream gauge location.

Upon receiving a river gauge forecast $L$, one may retrieve the simulation whose water level at the gauge location most closely matches $L$, thus obtaining a flood extent forecast while circumventing the need for a discharge measurement.\footnote{In effect, this procedure incorporates the estimation of a rating curve---the function relating water level to discharge---into the simulation.} To account for measurement uncertainty and forecasting errors, we add normally-distributed random noise to the forecast water level. This induces a probability distribution over simulated flood extents, which can be interpreted as an inundation probability value for each grid cell. These probabilities are discretized to three levels and rendered to users as three regions, respectively marked Some Risk, Higher Risk, and Highest Risk.

\subsection{Computational complexity}

A major challenge in hydraulic modeling is computational complexity. We are implementing several techniques to address this challenge. One approach is parallelization of the PDE solver through domain decomposition \cite{smith1997domain}. But even on multiple CPU cores, processing time quickly becomes prohibitive as the resolution and coverage area increase.

In recent years, a new class of AI accelerators has emerged, primarily to target large-scale deep learning. Google's Cloud Tensor Processing Unit (TPU) \cite{Jouppi2017} is one such example. 
The Cloud TPU v3 offers $420 \times 10^{12}$ floating-point operations per second (FLOPS) and 128 GB of high bandwidth memory (HBM), and multiple cores can be connected through a high-speed toroidal mesh network into a ``pod'' that achieves 100+ peta-FLOPS and 32TB of HBM\@.

Despite having originally been designed specifically for deep learning, TPUs have successfully been applied to non-ML tasks that conventionally require specialized high-performance computing resources \cite{tpu_ising_sc19}. TPU pods constitute a class of interconnected accelerators particularly well-suited for scaling to large-scale, high-resolution grids.

To utilize these capabilities, we designed a TensorFlow PDE solver which facilitates programming distributed algorithms on TPUs in easy-to-express Python code. The finite difference computation is distributed across multiple cores following the SIMD paradigm to perform domain decomposition \cite{smith1997domain}. Finite difference computation is efficient owing to the finite difference kernel being expressed as a 1D convolution, which is computed using TPU matrix multiplier unit (MXU) passes. The current implementation gains an 85x speed-up using a single TPU core relative to our reference single CPU core implementation. 

We are also exploring the potential of blending ML algorithms into traditional numerical models. For example, recent work showed that modeling features with neural networks can allow for solving wave-like PDEs on much coarser grids than is possible with standard numerical methods \cite{bar2018data}.

\subsection{Predictive inundation model} \label{predictive}

ML can complement the hydraulic simulation approach described above, by training a model on past flood events to directly infer the inundation extent from water level measurements. To this end, we used satellite imagery, specifically the synthetic aperture radar ground range detected (SAR GRD) data from the Sentinel-1 satellite, from which we determine the inundation extent at known timepoints \cite{schumann2015microwave}. At any given region, a SAR image is available once every several days. We matched this data with historical water level measurements, and restricted attention to measurements exceeding a predefined flood warning level. This resulted in a total of 10--40 SAR measurements per location, based on data available since 2014, the year Sentinel-1 became operational.

Using this data, we trained a model to predict, for each pixel, whether it would be wet or dry given the value of the nearest water level measurement. We assumed for simplicity that any given pixel $(x,y)$ will be wet for measurement values above some threshold $T(x,y)$, and dry otherwise. The model is thus a binary classifier per pixel, and a threshold may be chosen from the historical training data to balance precision and recall. We chose two separate thresholds: a recall-oriented threshold generating a Some Risk region, and a precision-oriented threshold yielding a Highest Risk region. These predictions are combined with the hydraulic simulation, as described in Section~\ref{section:deployment}.

\subsection{Deployment in an operational setting}\label{section:deployment}

To deploy the warning system in a new region, several sources of data are required. First, satellite imagery is used to generate a high-resolution DEM (Section~\ref{dem_generation}). Next, we obtain real-time hydrological forecasts for a stream gauge in the given site: our current system receives water level forecasts from the Indian Central Water Commission. Finally, we collect observations of past flood extents from processed satellite SAR imagery \cite{schumann2015microwave}.

This data allows the construction of the hydraulic model (Section~\ref{section:hydraulic}) and the predictive inundation model (Section~\ref{predictive}). As these models tend to complement each other, they were combined as follows: The final Some Risk region was defined as the \emph{union} of the Some Risk regions of the two models, and the Highest Risk region was defined as the \emph{intersection} of the Highest Risk region of the two models. This tends to improve the performance relative to each independent model, without significantly changing the areas of the various risk regions.

In real time, upon receiving a river level measurement, we construct the corresponding risk map, and provide it to individuals and government agencies. To maximize penetration, individuals can view such warning maps through a number of Google products, including Android notifications to people in affected areas, Google Search results, and Google Maps.

\section{Metrics and results} \label{sec:metrics}

To measure the end-to-end accuracy of our system, we compare our forecast maps to the ground truth inundation maps for historical flood events, which are based on SAR imagery (see Section~\ref{predictive}). The forecast ``Some risk'' region is geared toward recall, while the ``Highest risk'' region is geared toward precision. To measure our accuracy in a manner that optimizes for this asymmetry, we use the following metrics:

\textbf{Some risk recall (SRR):} This is defined as the number of pixels correctly detected as wet in the ``Some risk'' region, divided by the total number of wet pixels in the ground truth.

\textbf{Highest risk precision (HRP):} This is defined as the number of pixels correctly detected as wet in the ``Highest risk'' region, divided by the total number of predicted wet in the ``Highest risk'' region.

\textbf{Risk area ratio (RAR):} This is defined as the ratio between the area of the ``Some risk'' region and the ``Highest risk'' region. This metric is an indicator of the uncertainty expressed by the differences between our regions.

\begin{figure}[ht]
\begin{subfigure}[b]{0.45\textwidth}
  \centering
  \includegraphics[width=\textwidth]{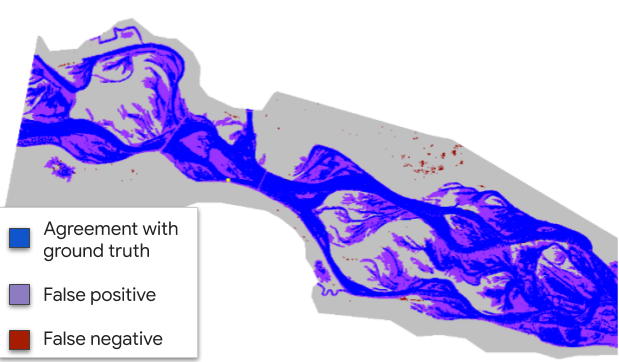} 
  \caption{Forecast for Patna for 2019-08-25, which achieved 99\% recall and 59.2\% precision.}
\end{subfigure}
\hspace{0.05\textwidth}
\begin{subfigure}[b]{0.45\textwidth}
  \centering
  \includegraphics[width=\textwidth]{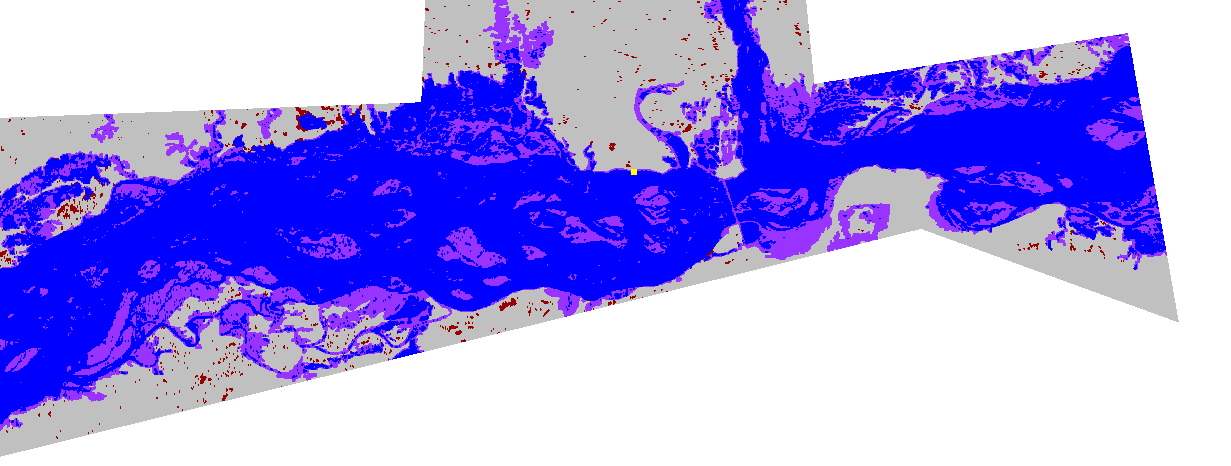} 
  \caption{Forecast for Tezpur for 2019-07-14, which achieved 98\% recall and 74.8\% precision.}
\end{subfigure}
\caption{Comparison of a ``Higher risk'' flood forecasts to the ground truth.}
\label{fig:tezpureval}
\end{figure}

Over the course of the 2019 monsoon season, our system achieved an average of 98.6\% SRR, 79.9\% HRP, and 2.81 RAR\@. Two typical forecast maps, both comparing the ``Higher risk'' regions with ground truth, are shown in Figure~\ref{fig:tezpureval}.

\subsubsection*{Acknowledgments}

This paper describes a large system which is the result of joint work of many people. We would like to particularly thank the following people for their contributions:
Jamie Adams,
Brett Allen,
John~Anderson,
Yi-Fan Chen,
Malo Denielou,
Mark Duchaineau,
Ran El-Yaniv,
Anton Geraschenko,
Pete Giencke,
Yotam Gigi,
Aleksey Golovinskiy,
Avinatan Hassidim,
Zhuoliang Kang,
Adi Mano,
Yossi Matias,
Paul Merrill,
Damien Pierce,
Slava Salasin,
Guy Shalev,
Rhett Stucki,
Ajai Tirumali,
and
Oleg Zlydenko.

\bibliography{bib_neurips_2019}
\bibliographystyle{plain}

\end{document}